\documentclass[letterpaper, 10 pt, conference]{ieeeconf}
\usepackage[bookmarks=false,hidelinks=true]{hyperref}
\usepackage{times}
\usepackage{graphicx,dblfloatfix}
\usepackage{mathrsfs,amsmath}
\usepackage{amssymb}
\usepackage{mathtools}
 
\usepackage{float}
\usepackage{listings}
\usepackage{color}
\usepackage{xcolor}
\usepackage{caption}
\usepackage{subfigure}
\usepackage{algorithm}
\usepackage{algpseudocode}
\usepackage{epigraph}
\usepackage{booktabs}
\usepackage{setspace}
\usepackage{amsfonts,amssymb}
\usepackage{multirow}
\usepackage{cuted}

\usepackage{tabularx}
\newcolumntype{L}[1]{>{\raggedright\arraybackslash}p{#1}}
\newcolumntype{C}[1]{>{\centering\arraybackslash}p{#1}}
\newcolumntype{R}[1]{>{\raggedleft\arraybackslash}p{#1}}

\IEEEoverridecommandlockouts
\title{\LARGE \bf
\emph{NDT-Transformer}: 
Large-Scale 3D Point Cloud Localisation \\
using the Normal Distribution Transform Representation}

\author{Zhicheng Zhou$^{1\dagger}$, Cheng Zhao$^{2\dagger}$, Daniel Adolfsson$^{3}$, Songzhi Su$^{4}$,\\
 Yang Gao$^{5}$, Tom Duckett$^{6}$, Li Sun$^{1*}$
  \thanks{$\dagger$Authors contribute equally.} 
  \thanks{This project is funded by EPSRC FAIR-SPACE Hub (EP/R026092/1), Royal Society RGS/R2/202432 and EU Horizon 2020 ILIAD (No. 732737).}
  \thanks{$*$The corresponding author. {\tt\small li.sun@sheffield.ac.uk}}
  \thanks{$^{1}$Visual Computing Group, University of Sheffield, UK}%
  \thanks{$^{2}$Department of Engineering Science, University of Oxford, UK}%
  \thanks{$^{3}$AASS Mobile Robotics Lab, Orebro University, Sweden}
  \thanks{$^{4}$IMT Lab, Xiamen University, China}
  \thanks{$^{5}$STAR LAB, Surrey Space Centre, University of Surrey, UK}
  \thanks{$^{6}$Lincoln Centre for Autonomous Systems, University of Lincoln, UK}
}

\begin{document}

\maketitle
\thispagestyle{empty}
\pagestyle{empty}

\begin{abstract}
3D point cloud-based place recognition is highly demanded by autonomous driving in GPS-challenged environments and serves as an essential component (i.e. loop-closure detection) in lidar-based SLAM systems.
This paper proposes a novel approach, named NDT-Transformer, for real-time and large-scale place recognition using 3D point clouds.
Specifically, a 3D Normal Distribution Transform~(NDT) representation is employed to condense the raw, dense 3D point cloud as probabilistic distributions (NDT cells) to provide 
the geometrical shape description. 
Then a novel NDT-Transformer network 
learns a global descriptor from a set of 3D NDT cell representations.  
Benefiting from the NDT representation and NDT-Transformer network, the learned global descriptors are enriched with both geometrical and contextual information. 
Finally, descriptor retrieval is achieved using a query-database for place recognition.
Compared to the state-of-the-art methods, the proposed approach achieves an improvement of 7.52\% on average top 1 recall and 2.73\% on average top 1\% recall on the Oxford Robotcar benchmark.
\end{abstract}
\section{INTRODUCTION}
\label{sec:introduction}
There is a high demand for point-cloud-based navigation systems that are able to robustly localise robots or autonomously driving cars in GPS-challenged environments. Point-cloud-based global localisation can also be employed as a loop-closure detection module, which is an essential component of Simultaneous Localisation And Mapping (SLAM) systems.
A practical method is to use GPS to acquire the coarse global location and point cloud registration methods, such as ICP, to obtain a more accurate pose estimate. However, GPS is not always available and reliable, so that alternative solutions using only sensory data are required.

\begin{figure}[thpb]
  \centering
    \includegraphics[width=0.99\linewidth]{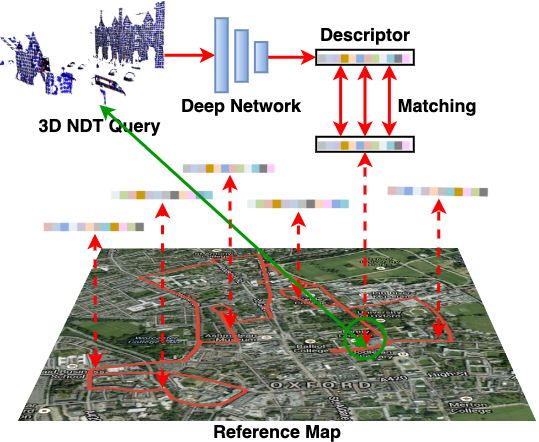}
    \caption{3D point~(NDT)-based loop-closure detection in a large-scale environment~(Oxford city centre).}
   \label{fig:overall}
\end{figure}


The conventional point cloud registration methods and particle filtering methods are not scalable for large-scale environments. 
The core challenges for current robot localisation systems
are generality and scalability. For example, model-based methods \cite{wang2020pointloc, ndt-gp,akai2020hybrid} can learn the 6\,DOF global pose as a regression model and infer the pose in real-time. Sadly, regression-based methods cannot be generalised to novel environments. SegMatch-like methods \cite{dube2017segmatch,segmap2018,tinchev2019learning,kong2020semantic} register two point cloud maps on intermediate representations (e.g.\ semantic segments), which has proven to be robust but not computationally efficient.


This paper addresses large-scale localisation as a large-scale point cloud retrieval problem, where the continuous robocar trajectories are discretized into consecutive `places'. Inspired by the success of the Transformer model~\cite{vaswani2017attention} in natural language processing, we propose a novel approach -- \emph{NDT-Transformer} -- to transform the point cloud into a `long sentence' of geometry-enriched NDT cells and then to a site-specific feature signature. During the inference, only the descriptor of the query data is required to be computed online, while the other descriptors in the database can be computed once offline and stored in memory, thus enabling real-time topological localisation in large-scale environments. Consequently, both scalability and generalisation can be achieved.

The main contributions can be summarised as follows:  
1) A computationally efficient method is proposed for large-scale point-cloud-based localisation where 
NDT is used as an intermediate representation. Our approach condenses a dense point cloud into a lightweight representation with maximal preservation
of the geometrical features;
2) A novel neural network architecture, named \emph{NDT-Transformer} is devised to learn a global descriptor with contextual clues from a set of 3D NDT cell representations;
3) The proposed method achieves the state-of-the-art performance in point-cloud-based place recognition, which can be an important supplement for NDT-based SLAM and Monte-Carlo localisation methods.

The processed data, code and trained model can be found on our project website: \url{https://github.com/dachengxiaocheng/NDT-Transformer.git}
\section{RELATED WORK}
\subsection{Point cloud-based global localisation}

The state-of-the-art point cloud based localisation methods can be categorised into three main streams: regression-based \cite{wang2020pointloc, ndt-gp,akai2020hybrid}, intermediate-representation-based
\cite{dube2017segmatch,segmap2018,tinchev2019learning} and global-feature-based \cite{he2016m2dp,kim2018scan,cop2018delight,kim20191,chen2020overlapnet,Arandjelovic_2016_CVPR,liu2019lpd}.  

Following the success of deep pose estimation in image-based global localisation, some methods \cite{wang2020pointloc, ndt-gp,akai2020hybrid} propose to use deep regression networks to learn the 6\,DOF global pose. These model-based approaches are extremely efficient but not generalisable. In other words, a new model is required to be trained for new environments. 

Intermediate-representation-based methods \cite{dube2017segmatch,segmap2018,tinchev2019learning,kong2020semantic} first segment the map into intermediate representations (i.e. semantic segments) then register the segmented parts, rather than directly registering the point clouds.
The intermediate parts can be described as distinctive features \cite{dube2017segmatch} or deep-learned features \cite{segmap2018,tinchev2019learning, kong2020semantic} may be used to eliminate the false positives. Though the matching efficiency can be improved by using intermediate representations, significant running time is spent on cloud segmentation, and thereby, real-time performance cannot be guaranteed. 

The mainstream methods for point-cloud-based localisation are based on global features, where a generalisable feature extractor is designed or learned to obtain the place signature for retrieval. Before deep learning began dominating the machine learning community, handcrafted features such as M2DP~\cite{he2016m2dp}, Scan Context~\cite{kim2018scan} and DELIGHT~\cite{cop2018delight} were well-studied to represent the 3D point cloud for the localisation.
In terms of learning-based methods, the Scan Context Image-based network~\cite{kim20191} and OverlapNet~\cite{chen2020overlapnet} convert the 3D lidar scan to a 2D image according to geometric knowledge, then deploy a 2D convolution-like network to learn a representation.


Instead of using hand-crafted features, PointNetVLAD~\cite{angelina2018pointnetvlad} combines PointNet~\cite{pntnet} and NetVLAD~\cite{Arandjelovic_2016_CVPR} to learn a global descriptor based on metric learning. 
However, the PointNet-like architecture ignores the spatial distribution and contextual cues within the 3D point cloud. 
Extracting efficient contextual information from the irregular 3D point cloud is another challenge for 3D loop-closure detection. 
The following work LPD-Net~\cite{liu2019lpd} employs a classical DGCNN~\cite{wang2019dynamic}-like network to enhance the feature descriptor by KNN-based aggregation in both feature space and Cartesian space.
They also introduce ten different kinds of geometric local features to feed the network during training, achieving the state-of-the-art~(SOTA) performance.
Most of these point-cloud-based methods require a downsampling operation on the original 3D dense point cloud due to the limitation of GPU memory.
However, this downsampling operation can cause a degeneration of the local geometric information.     
Lastly, as part of the real-time SLAM system, the run-time performance of 3D loop-closure detection needs to run in real-time, which means that lightweight networks are required.

\subsection{NDT-based localisation}
The Normal Distribution Transform (NDT) is a classic method to represent a 2D or 3D point cloud as differential multi-variant Gaussian distributions for 2D laser \cite{biber2003normal} and 3D lidar mapping \cite{stoyanov2013normal,magnusson2007scan}. Compared to point cloud or grid-map based representations, NDT-based relocalisation, i.e. NDT-MCL \cite{saarinen2013normal}, demonstrates advanced localisation precision and repeatability, as the NDT is an inherently probabilistic and geometrical representation to model the likelihood. In other words, the structural and geometrical information of the map is implicitly interpreted by the NDT cell parameters. Benefiting from this characteristic, NDT-based localisation can condense the bulky point cloud map to memory efficient NDT cells to make the localisation scalable. Submap-based NDT \cite{adolfsson2019submap} can eliminate the uncertainties caused by different robot perspectives, and further improve the localisation accuracy and scalability in large-scale and long-term applications \cite{ndt-gp}. Moreover, NDT can be used to provide place signatures for loop closure detection, e.g. using NDT histograms \cite{magnusspn_automatic_appearance} and semantic-NDT histograms \cite{zaganidis2019semantically},
semi-supervised place categorisation \cite{8202216}, and to create interest point descriptors \cite{7353812} for efficient registration or localisation.

\section{METHODOLOGY}
\label{sec:methodology}

\subsection{Problem Formulation}

Given a 6\,DOF trajectory of the robotcar and the synchronised lidar scans, submaps $M = \left \{ m_1, \cdots, m_N \right \}$ can be built by dividing the trajectory into fixed range intervals and mapping the point cloud into a local reference frame. 
Then the obtained submap can be further represented as NDT cells. 
Specifically, the space of a submap will be uniformly divided into grid cells, and the 3D points within a cell will be used to estimate the NDT parameters. 
Practically, the number of cells varies greatly due to the different density of each submap. 
A spatially distributed sampling filter $\mathcal{G}$ is applied to guarantee that the numbers of cells in all submaps are the same, i.e. $\left | \mathcal{G}(m_1) \right | = \cdots \left | \mathcal{G}(m_N) \right |$. A deep model called NDT-Transformer is proposed to learn a function $f(.)$ which represents the input NDT cell representations as 
a fixed-size global descriptor $f(\mathcal{F})$ where $\mathcal{F}= \mathcal{G}(m) $. 
Then the Euclidean distance function $d(.)$ can be adopted to measure the similarity, i.e. $d(f(\mathcal{F}), f(\mathcal{F}_{pos})) < d(f(\mathcal{F}), f(\mathcal{F}_{neg}))$, where $\mathcal{F}$ is similar to $\mathcal{F}_{pos}$ and dissimilar to $\mathcal{F}_{neg}$. 

Hence the problem of 3D loop closure detection is formulated as follows. Denoting the query submap as $m_q$, the task aims at searching the database $M$ for the most similar submap $m_*$. The problem can be resolved by searching for the nearest neighbour of a submap $m_* \in M$ in the feature space, i.e. $f(\mathcal{F}_{{m}_*})$, which has the smallest Euclidean distance to the query point cloud (feature) $f(\mathcal{F}_{{m}_q})$. That is, $d(f(\mathcal{F}_{{m}_q}), f(\mathcal{F}_{{m}_*})) < d(f(\mathcal{F}_{{m}_q}), f(\mathcal{F}_{{m}_i}))$, $\forall m_i \in M, i \neq * $. 

An example of 3D place recognition is illustrated in Fig.~\ref{fig:overall}. 
The red line on the map shows the entire route navigated by an autonomous vehicle in the city centre of Oxford, UK. 
A set of point clouds for this run can be constructed by combining the scans with global poses. 
The objective is to identify the nearest match of the query point cloud from another run. 
This paper represents all point clouds as 
a fixed number of 3D NDT cells (as shown in Fig \ref{fig:ndt}) and utilises the NDT-Transformer to convert them to site-specific global descriptors. 
Moreover, only the query descriptor is required to be computed online, while the other descriptors in the database are computed once offline and stored in memory, to guarantee the real-time performance.

\subsection{3D NDT Representation}
\begin{figure}[ht]
\centering
\includegraphics[width=0.99\linewidth]{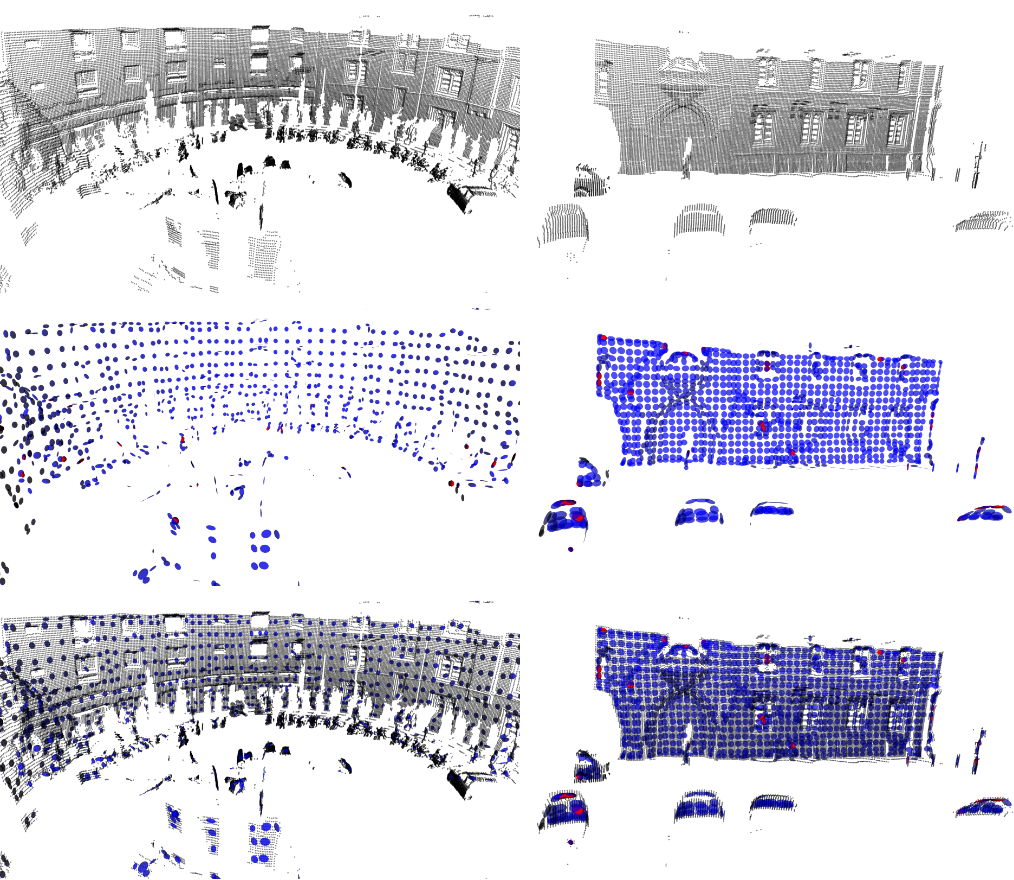}
\caption{First row: dense 3D point cloud, second row: 3D NDT cells, third row: 3D NDT cells within the dense 3D point cloud.}
\label{fig:ndt}
\end{figure}

In contrast to most of the existing work, we represent the point cloud submap using the 3D Normal Distributions Transform (NDT)~\cite{magnusson2007scan} rather than downsampling the 3D point cloud to a fixed number of points.
As illustrated in Fig.~\ref{fig:ndt}, the 3D-NDT representation is a compact `spherical' structure, which can not only preserve the geometrical shape of a group of local 3D points but can also significantly decrease the memory complexity. In other words, we can use memory efficient resolutions such as (1m $\times$ 1m $\times$ 1m) without losing significant geometrical details. The normal distribution $N(\mathbf{\mu}, \mathbf{C})$ of each NDT cell consists of a mean vector $\mathbf{\mu}$ and a covariance matrix $\mathbf{C}$ defined as,  

\begin{equation}
(\mathbf{\mu}, \mathbf{C})=(\frac{1}{n} \sum_{k=1}^n \mathbf{x}_k, ~ \frac{1}{n-1} \sum_{k=1}^n (\mathbf{x}_k - \mathbf{\mu}) (\mathbf{x}_k - \mathbf{\mu})^T).
\label{eq:1}
\end{equation}
where $\mathbf{x}_{k=1,\cdots,n}$ are 3D points in each cell. 



We employ a spatially distributed sampling approach $\mathcal{G}$ to generate evenly distributed NDT cells,
\begin{equation} 
\mathcal{F}:m^s \to \mathcal{G} ([\mathbf{\mu}_i, \mathbf{C}_i]_{i=1..k}),
\label{eq:3}
\end{equation}
where $m^s$ refers to submap of size $s$ and $\mathcal{G}$ refers to NDT representation of size $k$ cells.

The spatially distributed sampling can reduce the point density from an arbitrary input size $s$ to a fixed number of points $k$ and simultaneously preserve structural information from the original submap. This operation consists of three steps. Firstly, we downsample the original submap: $m^s$ of size $s$ to $m^{k*c\pm t\%}$ (within some tolerance factor $t$), where $c>1.0+t$ is an oversampling factor that makes sure that an excessive amount of points are produced. The downsampling is implemented by performing binary search to find the voxel size that produces $k*c\pm t$ points. Secondly, we estimate NDT distributions $\mathcal{F}$ with the downsampled points as centroids. That is, for each point in the downsampled submap $\mathbf{q}_i\in m^{k*c\pm t\%}$, an NDT cell $(\mu_i,\mathbf{C}_i)$ is computed from all points within a radius $r$ of $\mathbf{q}_i$ in the original submap $m^s$. The final step removes cells with the highest mutual information until $k$ cells remain. This ensures that the structural information is maintained. Mutual information is computed from the symmetric $Kullback–Leibler$ divergence $1/2*(D_{KL}(P||Q)+D_{KL}(Q||P) )$. The neighbour with the lowest $KL$-divergence has the highest mutual information. 

\begin{figure*}[t]
\centering
\includegraphics[width=\linewidth]{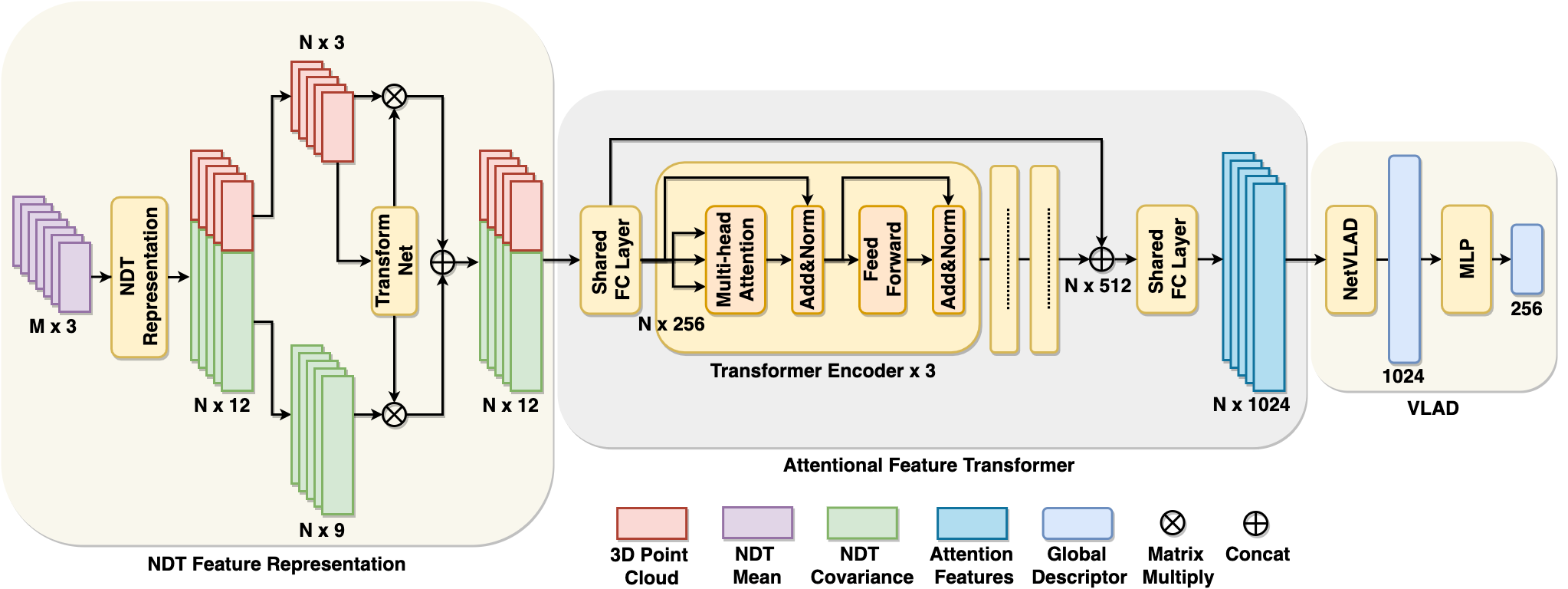}
\caption{NDT-Transformer Network Architecture}
\label{fig:network}
\vspace{-2ex}
\end{figure*}

\subsection{NDT-Transformer Network}
After converting the 3D submap $m$ to 3D NDT representation $\mathcal{F}$, this representation is fed into a NDT-Transformer network $f$ to obtain a descriptor $\xi$ of the 3D submap. That is, $\xi$\,=\,$f(\mathcal{F})$ where $\mathcal{F}$\,=\,$\mathcal{G}(m)$.  
As shown in Fig. \ref{fig:network}, the NDT-Transformer network consists mainly of three components: 1) NDT representation module that consists of a point transform and uncertainty backpropagation, 2) A residual transformer encoder, and 3) Net Vector of Locally Aggregated Descriptors~(NetVLAD).   

In order to achieve rotational invariance of the descriptor, the point transform and uncertainty backpropagation are applied to each original NDT representation $\mathcal{F}$\,=\,$(\mathbf{\mu}, \mathbf{C})$. 
Firstly, a transition matrix $T_{3 \times 3}$ is learned through a Transform Net~(T-Net)~\cite{pntnet} from the NDT mean vectors $\mathbf{\mu}$.
Then a new NDT representation $\mathcal{F}_{T}$\,=\,$(\mathbf{\mu}_{T}, \mathbf{C}_{T})$ can be obtained by assembling the transformed points and propagated uncertainties, 
\begin{equation} 
\mathcal{N}(\mathbf{\mu}_{T}, \mathbf{C}_{T}) = \mathcal{N}( T_{3 \times 3} \,\mathbf{\mu}, ~ T_{3 \times 3}\, \mathbf{C}\, T_{3 \times 3}^{T}),
\label{eq:4}
\end{equation}
where $T_{3 \times 3}$\,=\,$\mathcal{T} (\mathbf{\mu})$ and $\mathcal{T}$ refer to the T-Net.
Feature aggregation on irregular data such as NDT cells is still a challenging problem.  
PointNetVLAD~\cite{angelina2018pointnetvlad} utilises PointNet~\cite{pntnet} to transform the points' position information to a high-dimensional feature representation. 
However, this representation of each point suffers from the ambiguities due to a lack of contextual cues.  
Although local feature aggregation can be achieved via mean and covariance calculation within each NDT cell, there is no aggregation between the individual NDT representations. 
Our intuition is to employ an attention mechanism to learn the underlying context between landmarks (NDT cells).
In contrast to most existing approaches~\cite{liu2019lpd,wang2019dynamic}, which utilise kNN grouping for contextual information modelling, we employ a residual transformer encoder to aggregate the contextual cues for one NDT cell from the other NDT cells to increase the distinctiveness of the representation.

As shown in the middle panel of Fig. \ref{fig:network}, the network includes three transformer encoder stacks in series, and two shared linear stacks at the head and bottom with a shortcut skip connection.
Each Transformer encoder is composed of a series of modules, i.e. multi-head self-attention~(MHSA), feed-forward network~(FFN) and layer normalization~(LN), which can be stacked on top of each other multiple times.
The transformer encoder can learn co-contextual information/message $Attn$ captured by a self-attention mechanism,
\begin{equation} 
Attn([Q_i, K_i, V_i]) = \textrm{concat}([\textrm{softmax}(\frac{Q_i \cdot K_i^T}{\sqrt{d_k}})V_i]),
\label{eq:5}
\end{equation}
where $d_k$ denotes the dimension of queries and $Q_i, K_i, V_i$ stand for $i$th head of queries, keys, values of the NDT cell representation, respectively. 
In our implementation, four-head attention (i.e. $i={1,2,3,4}$) is used to enhance the discriminativeness of feature attributes.
The self-attention mechanism can automatically build connections between the current NDT cell and the other salient interesting NDT cells.
The attentional representation $\Phi$ can be obtained as, 
\begin{equation} 
\Phi' = LN ( \mathcal{F}_T + Attn ),
\label{eq:6.1}
\end{equation}
\vspace{-4mm}
\begin{equation} 
\Phi = LN ( FFN(\Phi' ) + \Phi'  ).
\label{eq:6.2}
\end{equation}
%
%
By this means, the local representation $\mathcal{F}_T$ of each NDT cell is upgraded to an attentional representation $\Phi$. 

Following PointNetVLAD~\cite{angelina2018pointnetvlad}, we choose NetVLAD~\cite{Arandjelovic_2016_CVPR} instead of a max-pooling layer to improve permutation invariance for the descriptor of the 3D point cloud. 
NetVLAD is designed to aggregate a set of local descriptors and generate one global descriptor vector.
It records statistical information with respect to local signatures and sums the differences between these signatures and their respective cluster.
In contrast to conventional VLAD, the parameters of NetVLAD, especially the assignment score, are learnt during training in an end-to-end manner. 
After going through the NetVLAD followed by a Multi-Layer Perceptron~(MLP), a set of NDT cell representations/descriptors $\Phi$ can be fused to obtain a fixed-size global descriptor vector $\xi$ to describe the 3D point cloud,
\begin{equation} 
\xi = MLP(NetVLAD(\mathcal{F}_T \oplus \Phi)),
\label{eq:7}
\end{equation}
 where $\oplus$ refers to the concatenation.

\subsection{Metric Learning}

Following the PointNetVLAD~\cite{angelina2018pointnetvlad}, the Lazy Quadruplet loss is used to achieve metric learning between the query global descriptor $\xi_{q}$ and the positive $\xi_{pos}$, the negative $\xi_{neg}$, and the hard negative $\xi_{neg*}$ examples, which are randomly picked 
during training 
from different locations to $\xi_{q}$. 
The Lazy Quadruplet loss is:
\begin{equation} 
\centering
\begin{split}
\mathcal{L}_{Q}(\xi_{q}, \xi_{pos}, \xi_{neg}, \xi_{neg*}) =
[ \alpha + d(\xi_{q}, \xi_{pos}) ~~~~~~~~~\\
- d(\xi_{q}, \xi_{neg}) ]_+ +  \left[ \beta+ d(\xi_{q}, \xi_{pos}) - d(\xi_{q}, \xi_{neg*}) \right ]_+,
\label{eq:8}
\end{split}
\end{equation}
where $\alpha$, $\beta$ are constant margin values, and $d(\cdot,\cdot)$ is the Euclidean distance.
The core idea of Quadruplet loss is to minimise the distance between the query and positive global descriptors, and maximise the distance between the query and negative global descriptors, while simultaneously keeping a proper distance between all negative descriptors (i.e. $\xi_{neg}$ and $\xi_{neg*}$).



\section{Experiment}

\subsection{Dataset Benchmark}
The Oxford Robotcar dataset~\cite{RobotCarDatasetIJRR}, which provides long-term multi-session data, was used to evaluate the proposed method.
In this 
dataset, a Sick LMS-151 2D lidar scanner mounted on the rear of the vehicle is used for mapping. 
Each scan contains a set of triplet records $(x, y, R)$, where the former two refer to coordinates in 2D Cartesian space and the latter refers to lidar infrared reflectance information. 
The dataset provides an accurate 6\,DOF trajectory of the travelled urban environment through fusing GPS-RTK with visual-inertial sensors, and a dense point cloud map can be built by transforming 2D lidar scans with respect to the global poses.    
Then places are generated
with a fixed interval (10\,m for training and 20\,m for evaluation) and the global 3D map is divided into a set of submaps using a fixed length trajectory segment (20\,m).

In this experiment, we use 44 runs for training and 22 runs for testing from the Oxford RobotCar dataset.
We obtain a total of 21898 submaps for training and 3071 submaps for testing.
For the NDT cell generation, we use 
2000 cells with a resolution of 0.8\,m as the input to 
We follow the definition of query, positive and negative data-pairs in PointNetVLAD~\cite{angelina2018pointnetvlad},  that is, the places within 10\,m to the query are classified as positives while places beyond 50\,m to the query are classified as negatives via KD-Tree search. 

\subsection{Neural Network Training}
Similar to PointNet~\cite{pntnet}, an input Transform Network (T-Net) is required to achieve rotational invariance by transforming the input map to a canonical view.
The hyperparameters of T-Net are the same as PointNet~\cite{pntnet}.
The two shared linear stacks before and after transformer encoders consists of linear layers with hidden variables with 256 and 1024 separately followed by Batch Normalisation layer and ReLU activation layer. 
The three Transformer encoders have the same hyperparameter settings: the dimensions of the input and output are set to 256, the dimension of the feed-forward layer is set to 1024, the dropout rate is 0.1, and the number of heads is set to 4. 
For the hyperparameter settings within NetVLAD, the dimensions of the input and output are set to 1024 and 256, respectively, and the size of the cluster is set to 64. 

For the metric learning, the network is trained with 20 epochs with a batch size of 2. 
There are 1 query submap, 2 positive submaps, 18 negative submaps, and 1 hard negative submap in one single batch.  
Margin values of $\alpha=0.5$ and $\beta=0.2$ are used.
The multi-step learning policy is used and the learning rate decay is a fixed value of $0.1$ applied on epoch 9 and 15. 
The initial learning rate is 1e-5 and the momentum is a fixed value of $0.9$.
The network is implemented in Pytorch and 
trained on a tower with Intel i9 CPU and two NVIDIA RTX Titan GPUs accelerated by CUDA and cuDNN.

\subsection{Performance Evaluation}

Following PointNetVLAD~\cite{angelina2018pointnetvlad} on the Oxford Robotcar dataset benchmark~\cite{RobotCarDatasetIJRR}, the recall indices, i.e., the Average Recall@N~(N=1,2,3...25) and Average Recall@1\% are employed as metrics for the performance evaluation of 3D loop closure detection. 

\begin{figure}[ht]
\centering
\includegraphics[width=0.99\linewidth]{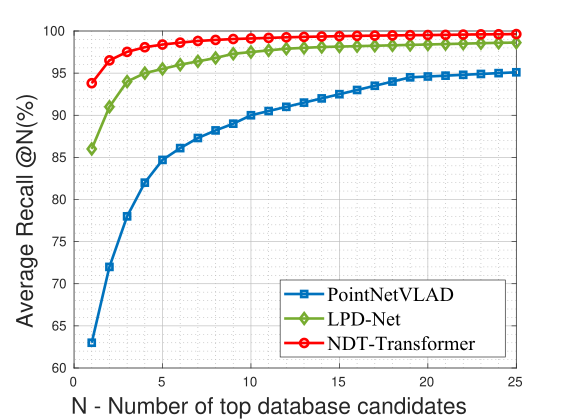}
\caption{The average recall curves from top 1~(@1) to top 25~(@25) of PointNetVLAD, LPD-Net and NDT-Transformer.}
\label{fig:comparison recall curves}
\vspace{-4ex}
\end{figure}

NDT-Transformer achieves $93.80\%$ top 1 (@1) average recall and $97.65\%$ top 1\% (@1\%) average recall, respectively.  
As shown in Table~\ref{table:comparison performance}, ND-Transformer achieves a significant improvement compared to PointNetVLAD~\cite{angelina2018pointnetvlad} with an improvement of $31.04\%$ on top 1(@1) average recall and an improvement of $16.64\%$ on top 1\% (@1\%) average recall. 
The performance of NDT-Transformer is superior to the SOTA, LPD-Net~\cite{liu2019lpd}, with $7.52\%$ improvement for top 1(@1) average recall and $2.73\%$ improvement for top 1\% (@1\%) average recall.
Although LPD-Net utilises 10 different kinds of local features, the NDT-Transformer only uses the NDT representations.
The average recall curves from top 1~(@1) to top 25~(@25) of PointNetVLAD, LPD-Net and NDT-Transformer are provided in Fig~\ref{fig:comparison recall curves}.

The ablation studies for NDT-Transformer are provided in Table~\ref{table:ablation studies}.
The average recall of top 1 (@1) and top 1\% (@1\%) decrease $17.87\%$ and $9.76\%$, respectively, if only the point position without covariance matrices is used for learning. 
The average recall of top 1 (@1) and top 1\% (@1\%) show a decrease of $6.92\%$ and $3.2\%$, respectively, if only the covariance matrices (i.e. without point position) are used for the network training.
If the T-Net is excluded from the network, the average recall of top 1 (@1) and top 1\% (@1\%) decrease by $1.98\%$ and $1.05\%$ due to the absence of the point transform and uncertainty propagation for the NDT representation.  
Average recalls from top 1~(@1) to top 25~(@25) of the studied ablation baselines are provided in Fig.~\ref{fig:comparison recall curves}.

\begin{table}[h!]
\centering
\begin{tabular}{l |c| c} 
 Network & Ave recall @1 & Ave recall @1\%  \\
 \hline
 PointNetVLAD\cite{angelina2018pointnetvlad} & 62.76 & 81.01 \\
 LPD-Net~\cite{liu2019lpd} & 86.28 & 94.92  \\
 NDT-Transfomer & \textbf{93.80} & \textbf{97.65} \\
\end{tabular}
\caption{The comparison result of PointNetVLAD, LPD-Net and NDT-Transformer, where average recall~(\%) at top 1~(@1) and the top 1\%~(@1\%) are reported.
}
\label{table:comparison performance}
\vspace{-3ex}
\end{table}

\begin{table}[h!]
\centering
\begin{tabular}{l |c| c} 
 Network & Ave recall @1 & Ave recall @1\%  \\ 
 \hline
 NDT-Transfomer-P & 75.93 & 87.89 \\
 NDT-Transfomer-C & 86.88 & 94.45 \\
 NDT-Transfomer-noT & 91.82 & 96.60 \\
 NDT-Transfomer & \textbf{93.80} & \textbf{97.65} \\
\end{tabular}
\caption{The ablation study of NDT-Transformer is given.
-P refers to our method using 3D points only. -C refers to our method using covariance matrices only.
noT is the network architecture without using T-Net.
}
\label{table:ablation studies}
\end{table}

\begin{figure*}[t]
\centering
\includegraphics[width=0.245\linewidth]{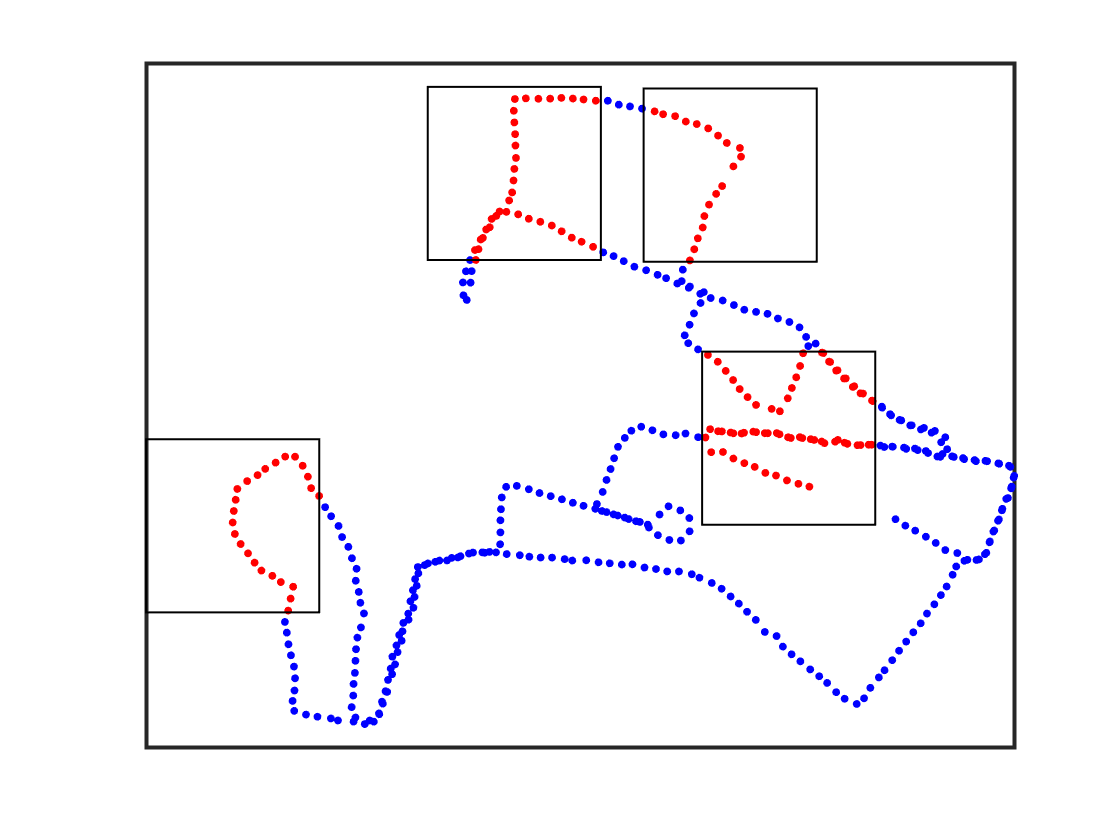}
\includegraphics[width=0.245\linewidth]{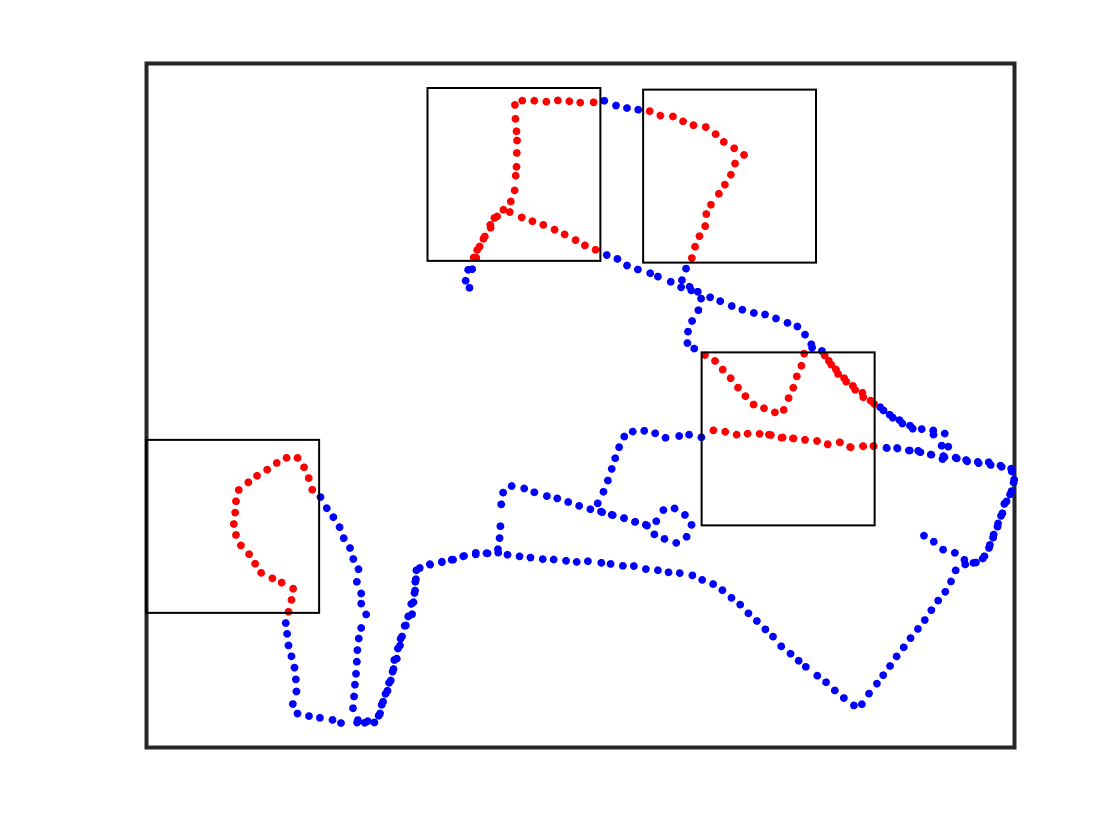}
\includegraphics[width=0.245\linewidth]{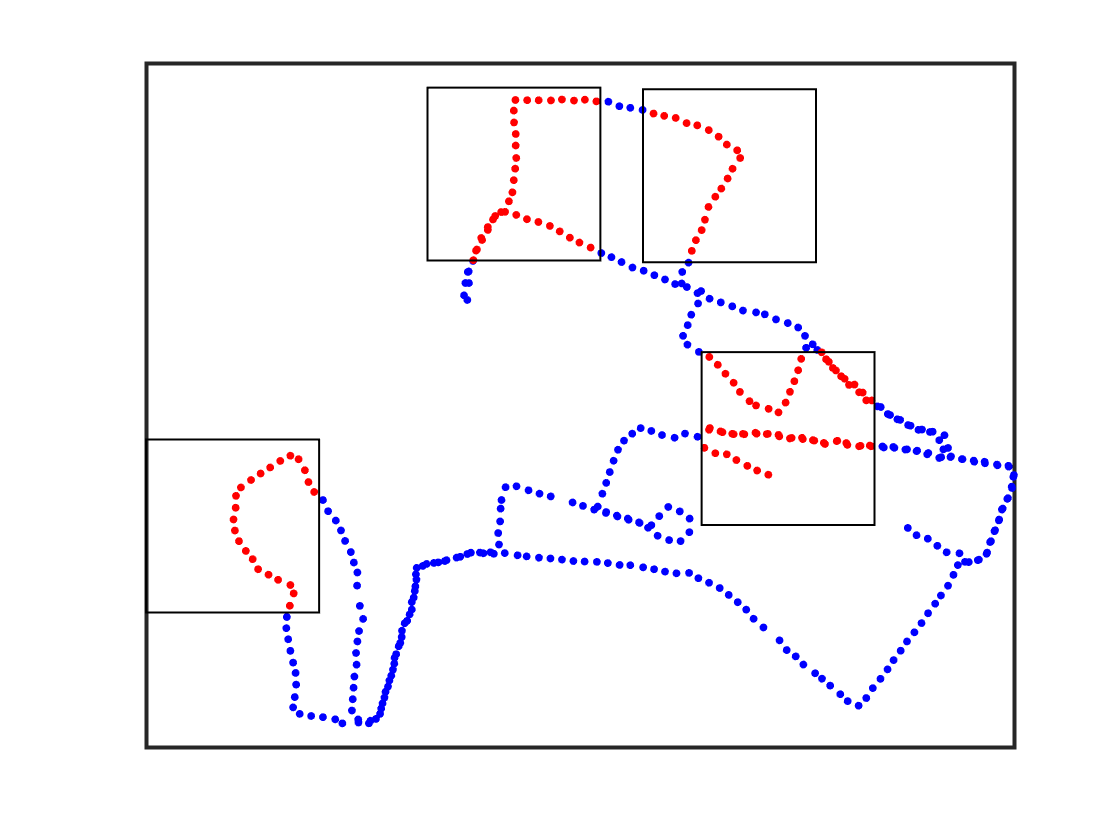}
\includegraphics[width=0.245\linewidth]{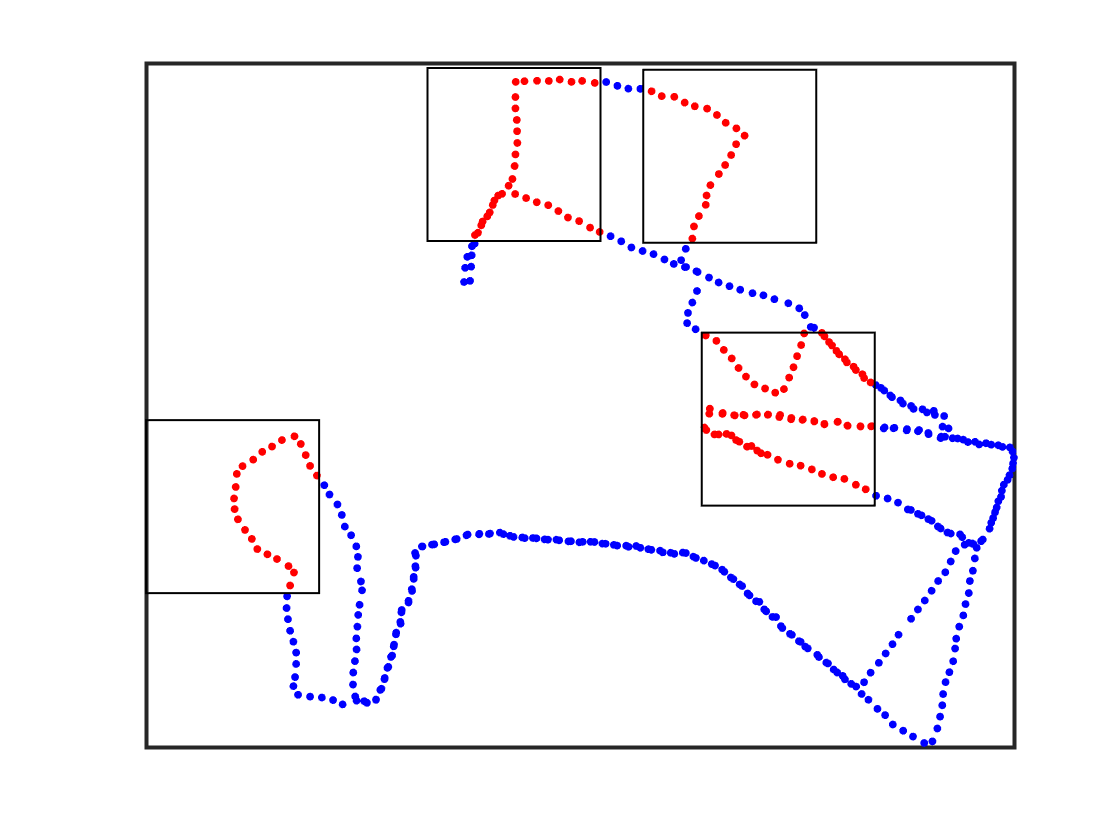}
\includegraphics[width=0.245\linewidth]{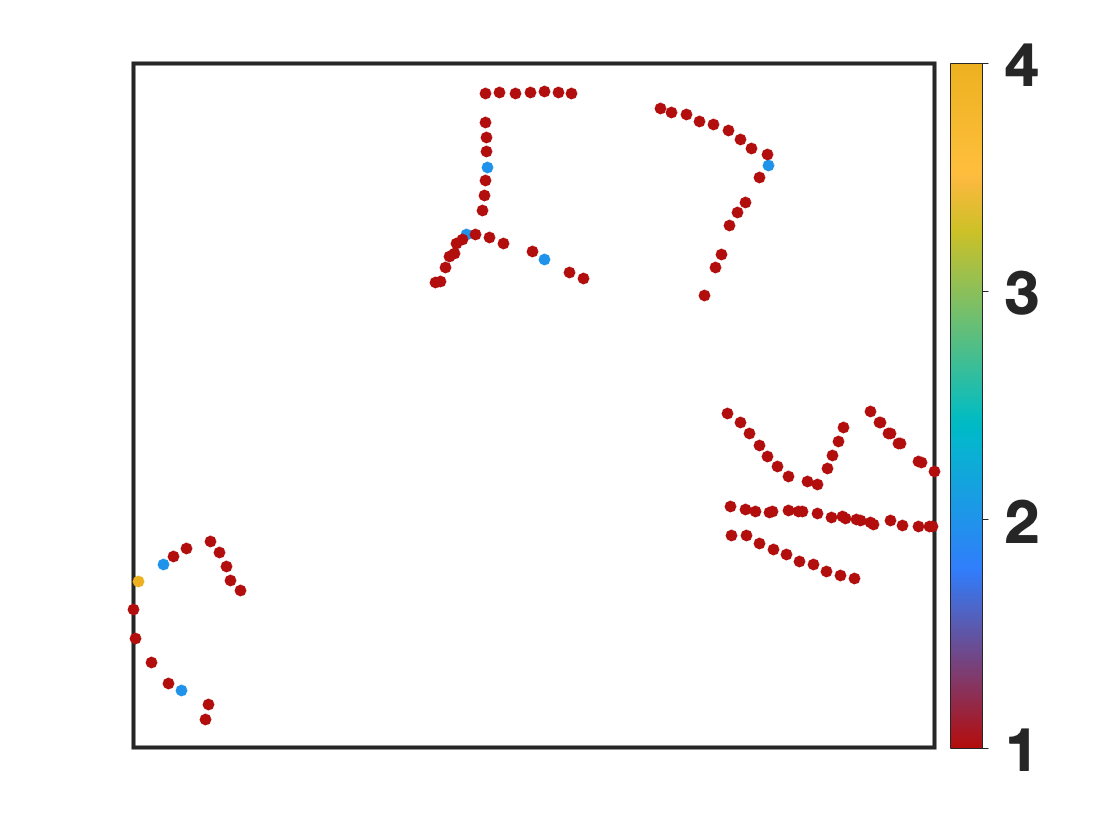}
\includegraphics[width=0.245\linewidth]{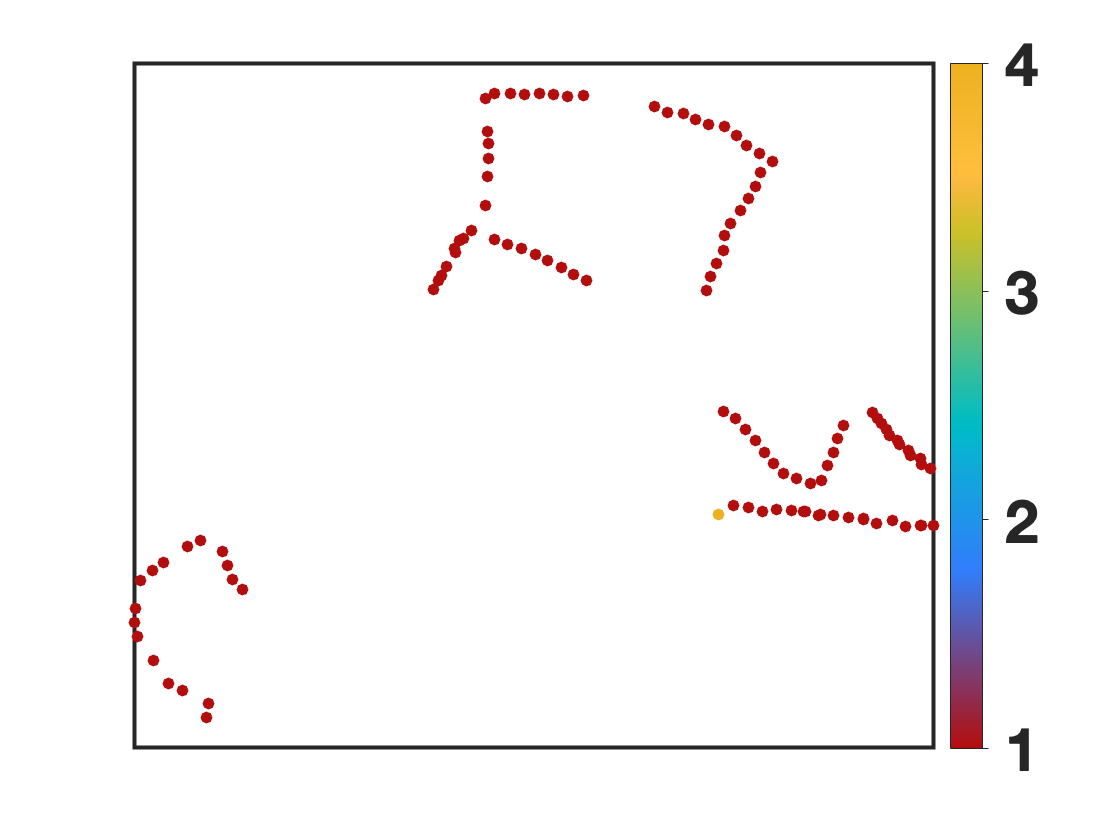}
\includegraphics[width=0.245\linewidth]{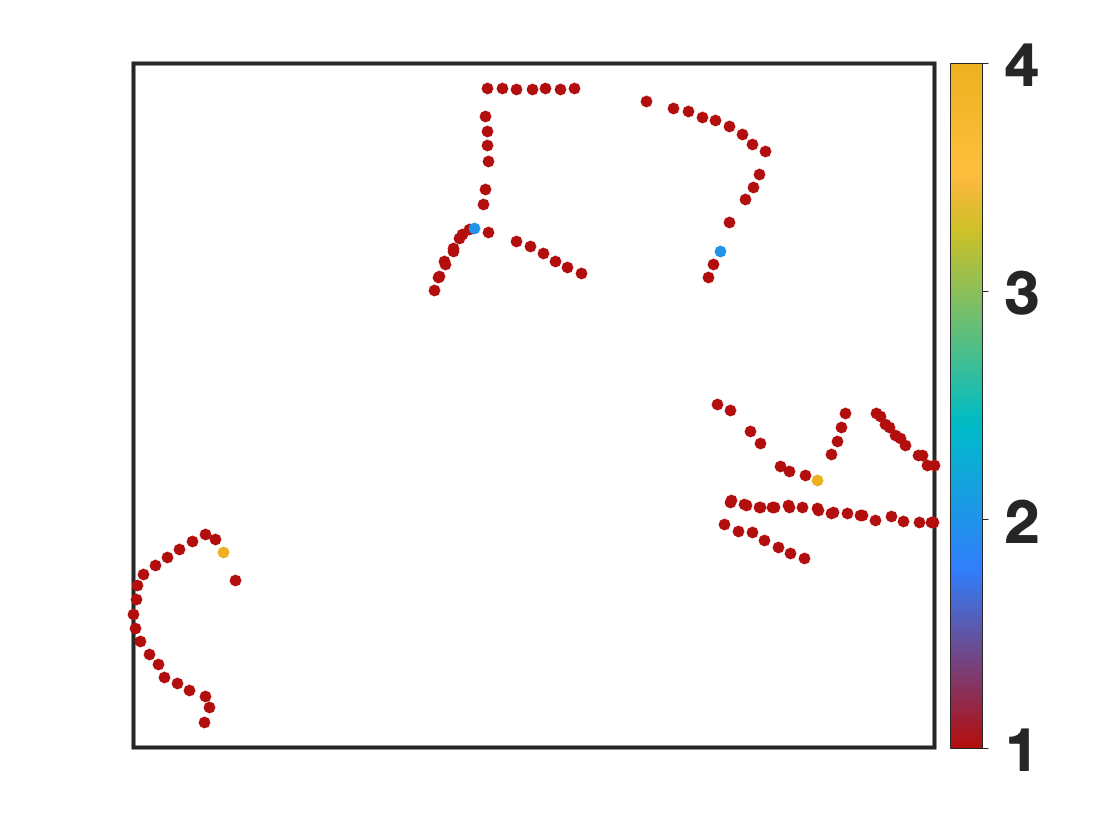}
\includegraphics[width=0.245\linewidth]{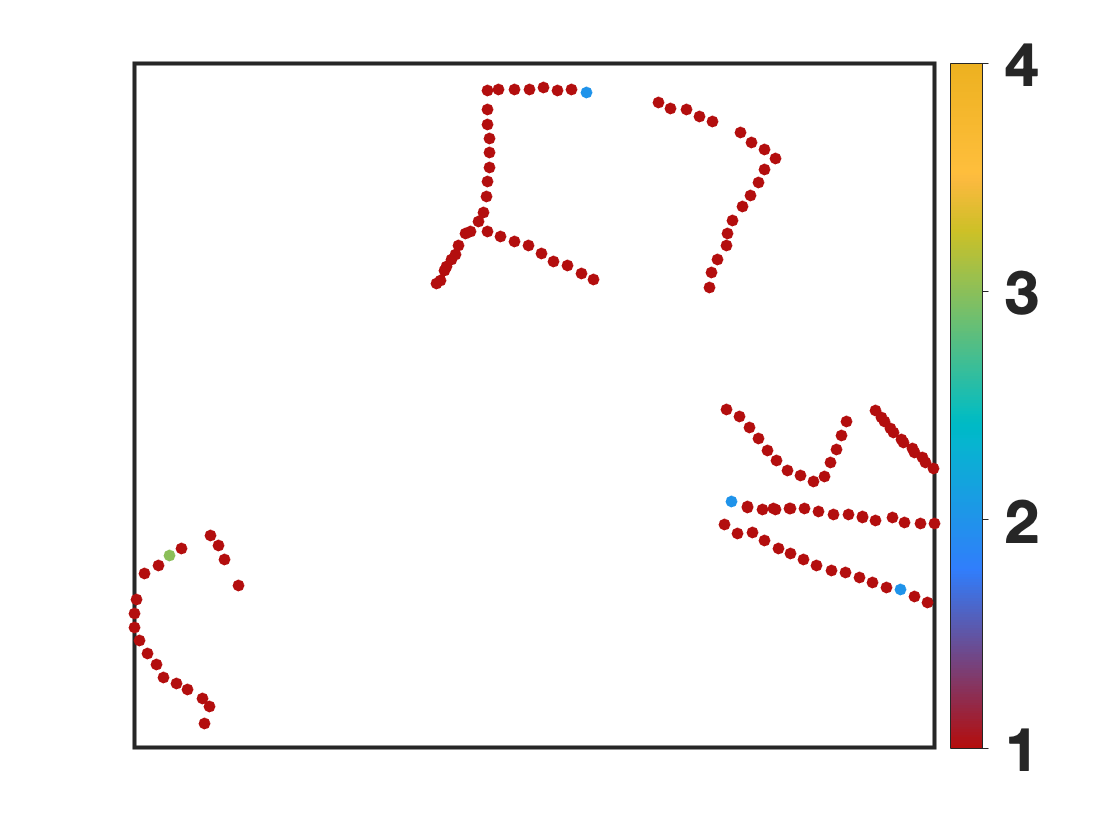}
\caption{In this figure, some examples of the database-query results of the proposed NDT-Transformer are shown.
Each column refers to an experiment randomly picked in our evaluation (i.e. date 2014-11-14, 2014-12-09, 2015-02-03, 2015-04-24, on Oxford Robocar dataset). The first row shows the database locations where red dots are testing areas (i.e. streets never seen before). The second row shows the retrieval results (the ranking of successfully retrieved candidates)}. 
\label{fig:retrieved colour map}
\vspace{-6ex}
\end{figure*}

\begin{figure}[ht]
\centering
\includegraphics[width=0.99\linewidth]{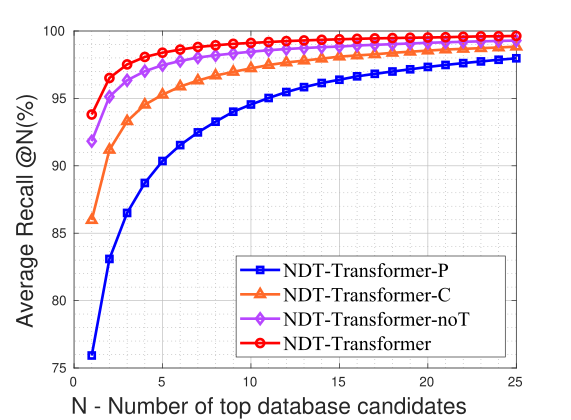}
\caption{The average recall curves from top 1~(@1) to top 25~(@25) of the NDT-Transformer ablation studies.}
\label{fig:ablation recall curves}
\vspace{-2ex}
\end{figure}

Finally, we show some examples of retrieval results on the Oxford Robotcar dataset using NDT-Transformer, where the ranking of correctly retrieved candidates is plotted in different colours. As shown in Fig.~\ref{fig:retrieved colour map}, the first row images are database submaps randomly picked from 22 tests in the experiment, in which the red trajectories are queries from previously unseen areas. 
The second row shows the heat maps indicating the retrieval results. Colours from red to yellow represent correctly recognised places in the top 1 to 4 candidates.  
The KD-Tree is employed to search the top $N$ matched candidates from the database. The figure shows that most of the queries are paired with the correct database locations as the first ranking candidate. It worth noting that the query areas (shown as four rectangles) are previously unseen streets, which means that no data from these areas
were used in the training set.
Therefore, this experimental result shows that our deep-learned features can be used for localisation in unseen areas (novel locations). Moreover, the 
inference time of our approach averages 0.034 seconds per point cloud (0.03 seconds for NDT conversion and 0.004 seconds for the transformer network) and the computational complexity of retrieval is $\mathcal{O}(log n)$.


\section{Conclusion}
This paper proposes a novel approach, NDT-Transformer, to represent a dense point cloud submap as a discriminative location descriptor, to addresses the large-scale topological localisation problem by the means of place recognition and retrieval. Our approach leverages the generic NDT representation and transformer encoder to interpret the point cloud, where geometric and contextual information are implicitly summarised. Compared to the state-of-the-art method~\cite{liu2019lpd}, our approach does not require
handcrafted features, and instead aggregates and strengthens local features hierarchically, hence achieving better results with superior run-time performance. 

The experimental results on the Oxford Robocar dataset demonstrate the effectiveness of our approach, i.e. a significant improvement of $7.52\%$ on average recall of top\,1 and $2.73\%$ on top 1\%~ in comparison with the SOTA method. 
Our system is able to successfully find a compromise between the needs of SOTA performance and run-time performance as an essential module of a real-time SLAM system.
The proposed approach also provides an important supplement for NDT-based SLAM approaches such as NDT mapping, NDT-MCL, etc.



\bibliographystyle{IEEEtran}
{\bibliography{refs}}

\end{document}